\documentclass[runningheads]{llncs}

\usepackage{eccv}

\usepackage{eccvabbrv}

\usepackage{graphicx}
\usepackage{amssymb}
\usepackage{booktabs}
\usepackage{xcolor}
\usepackage{colortbl}
\usepackage{multirow}
\usepackage[dvipsnames]{xcolor}
\usepackage{wrapfig}
\usepackage{enumitem}
\usepackage{tabularx}

\definecolor{aliceblue}{rgb}{0.94, 0.97, 1.0}

\usepackage[accsupp]{axessibility}  

\usepackage{hyperref}

\usepackage{orcidlink}
\usepackage{fontawesome}

\begin{document}

\title{ToddlerAct: A Toddler Action Recognition Dataset for Gross Motor Development Assessment}

\titlerunning{ToddlerAct: A Toddler Action Recognition Dataset for Gross Motor Development Assessment}

\author{Hsiang-Wei Huang\inst{1}\orcidlink{0009-0009-2474-8869} \and
Jiacheng Sun\inst{1}\orcidlink{0009-0009-8710-7858} \and
Cheng-Yen Yang\inst{1}\orcidlink{0009-0004-2631-6756} \and 
Zhongyu Jiang\inst{1}\orcidlink{0000-0003-4462-6497} \\ 
Li-Yu Huang\inst{2}\orcidlink{0009-0008-2795-3270} \and
Jenq-Neng Hwang\inst{1}\orcidlink{0000-0002-8877-2421} \and
Yu-Ching Yeh\inst{2}\textsuperscript{\faEnvelopeO}\orcidlink{0000-0002-9530-6226}
\\ \textsuperscript{\faEnvelopeO} Corresponding Author}

\authorrunning{Huang et al.}
%
\institute{University of Washington, Seattle WA 98195, USA \\
\email{\{hwhuang,sjc042,cycyang,zyjiang,hwang\}@uw.edu}\\
\and
National Chiayi University, Chiayi City 60004, Taiwan \\
\email{\{s1105030,ycyeh\}@mail.ncyu.edu.tw}\\}

\maketitle

\begin{abstract}
Assessing gross motor development in toddlers is crucial for understanding their physical development and identifying potential developmental delays or disorders. However, existing datasets for action recognition primarily focus on adults, lacking the diversity and specificity required for accurate assessment in toddlers. In this paper, we present ToddlerAct, a toddler gross motor action recognition dataset, aiming to facilitate research in early childhood development. The dataset consists of video recordings capturing a variety of gross motor activities commonly observed in toddlers aged under three years old. We describe the data collection process, annotation methodology, and dataset characteristics. Furthermore, we benchmarked multiple state-of-the-art methods including image-based and skeleton-based action recognition methods on our datasets. Our findings highlight the importance of domain-specific datasets for accurate assessment of gross motor development in toddlers and lay the foundation for future research in this critical area. Our dataset will be available at \url{https://github.com/ipl-uw/ToddlerAct}.

\end{abstract}
\section{Introduction}
Early childhood development encompasses various domains, including cognitive, social, and physical development. Gross motor skills, which involve large muscle groups and whole-body movements, are fundamental for toddlers' physical development and overall well-being. Assessing gross motor development in toddlers is crucial for identifying potential delays or disorders early on, allowing for timely interventions and support. Traditional assessment methods often rely on subjective observations by clinicians, which can be time-consuming and prone to bias. To this reason, we seek the solution in using computer vision method to solve the problem.

With the recent development in computer vision field, more and more research started to focus on human-oriented understanding, including detection \cite{Jocher_Ultralytics_YOLO_2023}, pose estimation \cite{yang2022unsupervised,zhouran,ho2024rt,jiang2024back}, tracking \cite{zhang2022bytetrack,huang2024iterative,mttrack,huang2023enhancing} and action recognition \cite{infact,mazzia2022action}. Action recognition, is a crucial technology that can be applied to different real world applications like anomaly detection, fall/fight detection, action assessment...etc. Action recognition for toddler, is an important area of research with significant implications for early childhood development. The ability to accurately recognize and assess toddler actions can lead to more objective and efficient evaluations of gross motor skills, facilitating early detection of developmental delays or disorders. 

However, there are several challenges in toddler action recognition. Including the challenges in dataset collection, the variability of toddler movements, and the need for robust models capable of handling such variability. Collecting a high-quality dataset for toddler action recognition poses significant difficulties, mostly due to the ethical considerations and privacy concerns must be meticulously addressed when involving young children in research. Another challenge is the inherent variability in toddler movements. Unlike adults, toddlers exhibit a broader spectrum of motion patterns and speeds, which can be influenced by their developmental stage, mood, and environment. This variability makes it difficult to develop models that can generalize well across different individuals and scenarios. Lastly, a robust model are essential to accurately recognize and assess toddler actions. The current pretrained action recognition model usually focus on adult action recognition, the scarcity of data is also making toddler action recognition model to struggle with getting better performance.

To address these issues, we present a straightforward toddler action recognition baseline that leverages the foundation model CLIP \cite{radford2021learning} to conduct action recognition. To further advance research in toddler action recognition, we also collect a comprehensive dataset containing more than 500 video-action pairs and 70,000 frames, all annotated with expert-labeled action categories. 

In summary, this paper introduces several main contributions including:
\begin{itemize}
    \item[$\bullet$] Proposed a large scale toddler action recognition dataset, named ToddlerAct, containing 495 videos with over 500 tracklet-action pairs collected from real world and are annotated with tracklet and action label by multiple experts. The dataset will be released upon acceptance.
    \item[$\bullet$] Benchmarked action recognition method that leverages CLIP as the backbone to conduct action recognition and also multiple state-of-the-art action recognition methods to serve as baselines for our proposed dataset.
    \item[$\bullet$] Conduct experiments using our proposed method on the ToddlerAct dataset, discuss the challenge and the potential impact of the toddler action recognition in gross motor development assessment field.
\end{itemize}

\section{Related Work}
\subsection{Gross Motor Development Assessment}
Gross motor development assessment is a critical area in early childhood development, focusing on evaluating children's abilities to perform activities that involve large muscle groups and whole-body movements. Traditional methods of assessment include standardized tests like the Peabody Developmental Motor Scales and the Alberta Infant Motor Scale. These assessments often require trained professionals to observe and score children's performance on various tasks, which can introduce subjectivity and bias. Recent advancements in technology have led to the development of automated systems that use motion sensors, wearable devices, and computer vision techniques to provide more objective and efficient evaluations.

A longitudinal study \cite{toddler1} involving 36 infants indicated that gross motor skills and language development at 18 months can effectively predict spatial vocabulary ability at 3 years old. Infants' gross motor development can serve not only as an early assessment and diagnostic tool but also contribute to future child health policies, such as early head start program. The Multicentre Growth Reference Study (MGRS) promoted by WHO includes important indicators of gross motor development \cite{wijnhoven2004assessment}. The six gross motor milestones for infants before the age of 2 are: independent sitting, crawling, assisted standing, assisted walking, independent standing, and independent walking. Long-term longitudinal study results show that the timing of development of these six gross motor skills is significantly correlated with later fine motor skills, personal and social development, and overall developmental scores in young children \cite{toddler2}. When the MGRS study was conducted in 2004, data sources included assessments from both caregivers and trained MGRS fieldworkers. It was common for discrepancies to occur between their evaluations. When such discrepancies arose, the study relied on the observers’ judgment. Children’s health data collection required funding, and it is also time consuming. Lacking of internal consistency among observers was a potential issue in such studies \cite{toddler3}, and the assessment of infants' gross motor development often suffers from low consistency and accuracy due to the subjective judgment of observers. Additionally, because the evaluation of motor development frequently employs more than one tool, there could be discrepancies in the infant ages and specific movements measured by different tools  \cite{toddler4} (e.g., AIMS, DF-mot \cite{dfmot}, and Bayley Infant Scales \cite{bayley2006bayley}). Therefore, using computer vision to interpret infants' movements can provide relatively objective and consistent assessment results.

\begin{table*}[!t]
    \caption{An overview of the existing public child-related pose and action recognition datasets used in computer vision tasks. Real$^{*}$ represents the data are collected from the internet or other public available source while the rest are collected in a controlled environment.\\}
    \vspace{-12pt}
    \centering
    \footnotesize
    \begin{tabular*}{\textwidth}{@{\extracolsep{\fill}}lccccc}
        \toprule
        \textbf{Dataset}& \textbf{Year} &  \textbf{Videos} & \textbf{Frames} & \textbf{Source} & \textbf{Annotations}\\
        \midrule
        
        SyRIP \cite{syrip} & 2021 & - & 1,700 & Real$^{*}$ + Synthetic & 2D \& 3D Pose \\
        MINI-RGBD \cite{mini-rgbd} & 2018 &   12 & 12,000 & Synthetic & 2D \& 3D Pose \\
        BabyPose \cite{migliorelli2020babypose} & 2020 & 16 & 16,000 & Real & 2D Pose \\
        AggPose \cite{cao2022aggpose} & 2022 &  5,187 & 20,847  & Real & 2D Pose \\
        InfAct \cite{infact} & 2023  & 200 & 38,126 & Real$^{*}$ & Action \\
        MTTrack \cite{mttrack} & 2024  & 10 & 3,000 & Real & Tracklet \\
        \midrule
        ToddlerAct(Ours) & 2024  & 495 & 70,235 & Real & Tracklet \& Action \\
        
        \bottomrule
    \end{tabular*}
    \vspace{-18pt}
    \label{table:dataset}
\end{table*}

\subsection{Action Recognition Methods}
Action recognition is a well-established field in computer vision, with applications ranging from surveillance to human-computer interaction. With the advent of deep learning, convolution neural networks (CNNs) and recurrent neural networks (RNNs) have become predominant, allowing for end-to-end learning of features directly from raw data. Models like 3D CNNs \cite{feichtenhofer2019slowfast}, and transformer-based architectures \cite{mazzia2022action} have shown significant improvements in recognizing complex actions. Furthermore, some methods \cite{li2022uniformerv2} also leverage pretrained foundation model like CLIP \cite{radford2021learning} to conduct action recognition and achieve state-of-the-art performance. Beside image modality, some method include human keypoint as a way for action recognition, including ST-GCN \cite{stgcn} and PoseConv3D \cite{duan2022revisiting}, which also achieve decent performance on public dataset.

Most of the existing work focus on the action recognition of adult and daily behavior, while there are limited works focus on child action recognition, which is critical in child mental development assessment.

The existing action recognition benchmarks and methods mainly focus on infant action recognition, which only include child with age under one-year-old. However, the toddler's action recognition, is also important and even more complicated and challenging due to the capability of toddler to perform even more complex action compared to infant. To the best of our knowledge, there is no existing toddler action recognition dataset. To facilitate the related research in this field, we proposed the first toddler action recognition dataset, named ToddlerAct, aiming to enhance the research in toddler action recognition and development assessment research.

\begin{figure}[t]
    \centering
    \includegraphics[width=1\linewidth, height=200pt]{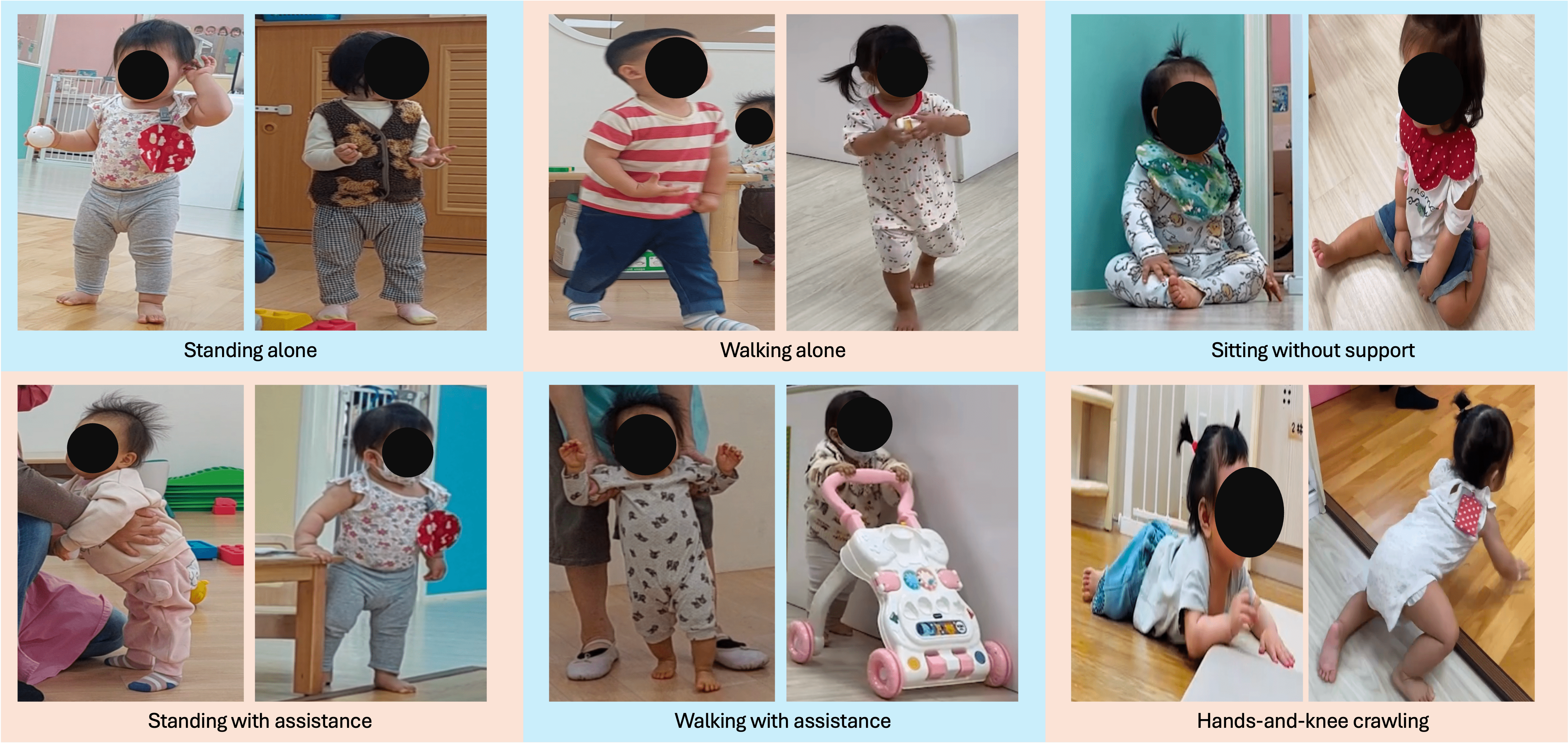}
    \caption{Some visualization samples of the six different actions contain in our dataset. All the toddlers' faces are masked to protect privacy in our dataset.}
    \label{fig:data}
\end{figure}

\subsection{Child Pose and Action Recognition Datasets}
The development of child-specific datasets is essential for advancing research in child pose estimation and action recognition \cite{syrip,yang2022unsupervised}. Existing datasets focusing on pose estimation include SyRIP \cite{syrip}, MINI-RGBD \cite{mini-rgbd}, BabyPose \cite{migliorelli2020babypose}, and AggPose \cite{cao2022aggpose}, while InfAct \cite{infact} focuses on the action recognition of infants. However, these existing datasets and methods primarily emphasize infant-related research, focusing on children under one year old. To our best knowledge, there is no existing dataset that focuses on toddler-related research, emphasizing children under three years old.

Research on toddlers should be given more attention due to the unique and rapid developmental changes occurring during this stage. Toddlers exhibit distinct and more complicated gross motor skills that differ significantly from those of infants, such as walking and crawling. These skills are crucial for their physical development and overall well-being. To address this gap, we propose the creation of a comprehensive toddler action recognition dataset, named ToddlerAct. This dataset include 495 videos with over 500 video-action pairs and 70,000 frames collected from the real world, all annotated with expert-labeled action categories following the WHO multicentre growth
reference study \cite{wijnhoven2004assessment}. By focusing on toddlers, this dataset aims to capture a wide range of activities and movements specific to this age group, providing a robust resource for researchers and practitioners. We hope this dataset will facilitate the development of advanced models for toddler action recognition, contributing to better early detection of developmental delays and enhancing the effectiveness of early interventions.

\begin{figure}[t]
    \centering
    \includegraphics[width=0.5\textwidth]{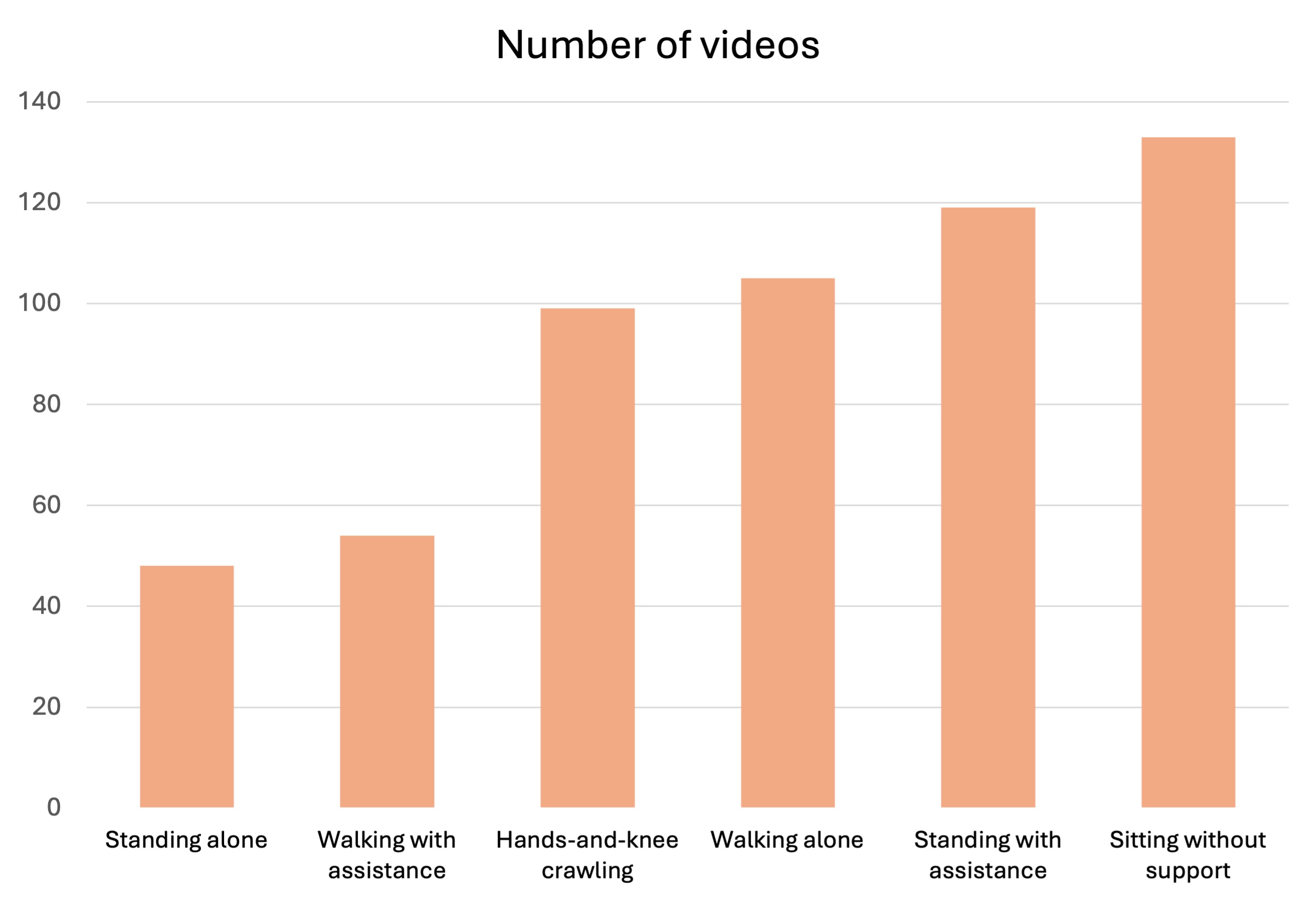}
    \caption{Number of videos per action in our ToddlerAct dataset.}
    \label{fig:numofsample}
\end{figure}

\section{Dataset and Baseline Methods}
\subsection{ToddlerAct Dataset}
Our proposed toddler action recognition dataset contains six different action annotations following the gross motor development in the WHO multicentre growth reference study \cite{wijnhoven2004assessment}. The six actions include sitting without support, hands-and-knees crawling, standing with assistance, walking with assistance, standing alone, and walking alone. Because a video might contain multiple toddlers performing different actions, we also conduct tracking using YOLOv8 \cite{Jocher_Ultralytics_YOLO_2023} and ByteTrack \cite{zhang2022bytetrack} to get bounding box and tracking ID. Finally, we conducted manual inspection and annotate the toddler's tracking ID and action pair for each video. The total data collection and annotation process took around 300 person-hours. The dataset contains 495 videos, with 556 tracklet-action pairs and 70,235 frames, annotated with tracklets (per frame bounding box and track ID) and the target toddler's action label for each video. To our best knowledge, the proposed dataset is currently the largest child-related research dataset in the computer vision field, exceeding previous datasets in terms of the total frame count and higher resolution.

\subsection{Image-based Method}
\subsubsection{Backbone.}
Due to the scarcity of data in toddler action recognition, large foundation model can serve as an effective way to conduct action recognition. We benchmarked a simple method using CLIP \cite{radford2021learning} to serve as one of the baseline for our dataset. CLIP is trained on billions of images using contrastive learning, which can be a strong backbone for our method to perform action recognition on the collected toddler action recognition dataset. CLIP stands for contrastive language image pretraining, which the architecture consists of an image encoder and a text encoder, given an image and text pair, CLIP maps the image and text into a shared embedding space, which can be further applied to various downstream tasks, including image classification, action recognition...etc.
In this baseline method, we use a pretrained CLIP image encoder to extract the visual feature from $n$ sampled frames from a video. The extracted feature are further send into a classification head to conduct action recognition and generate action label. This approach leverages the powerful feature extraction capabilities of CLIP and adapts them to the specific task of recognizing toddler actions. By doing so, we can effectively address the challenge of limited annotated data in this domain and establish a robust baseline for future research in toddler action recognition.

\begin{figure}[t]
    \centering
    \includegraphics[width=1\linewidth, height=170pt]{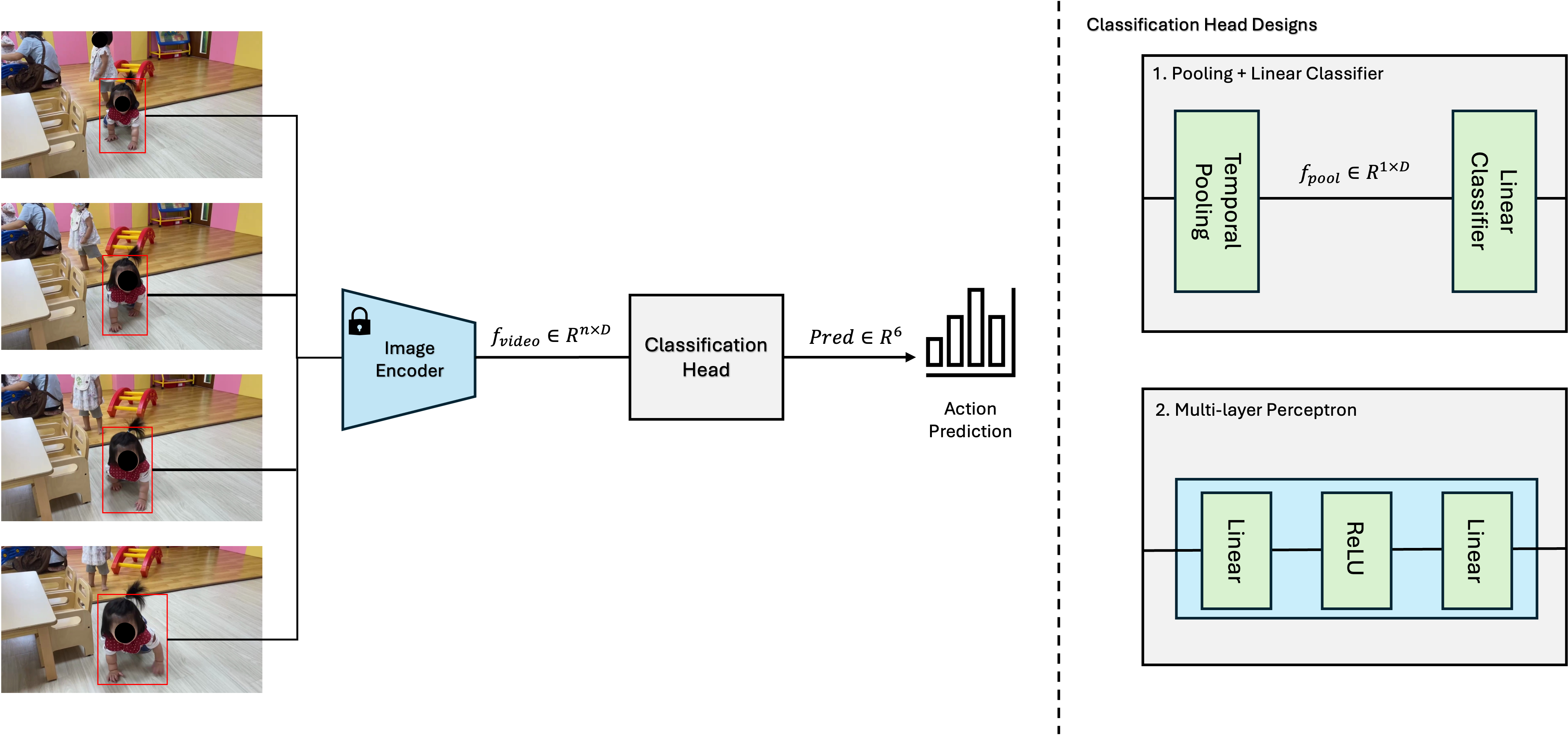}
    \caption{The overall pipeline of a simple CLIP-based action recognition baseline and the two different classification head designs. Both of the method utilized a frozen image encoder from CLIP. The first method incorporates a temporal pooling module (maximum pooling or average pooling) followed by a linear classifier, while the other incorporates a multi-layer perceptron to conduct action prediction.}
    \label{fig:method}
\end{figure}

\subsubsection{Classification Head.} After the visual features are extracted from each sampled frames $V_{i}$ in video $V$, we can obtain a series of sampled visual features $f_{video} = \{f_0, \dots, f_n\}$. With $n$ representing the number of sampled frames and $f_{video} \in R^{n \times D}$, $D$ is the hidden dimension of features extracted from CLIP. We further send the extracted features $f_{video}$ into a classification head, which we experimented with three different variants as shown in fig. \ref{fig:method}, including
\begin{enumerate}
    \item Average pooling of feature from $f_{video}$ followed by a linear classification layer.
    \item Maximum pooling of feature from $f_{video}$ followed by a linear classifier layer.
    \item A multi-layer perceptron that takes $f_{video}$ as input and generate action prediction.
\end{enumerate} 
The overall pipeline can be described in the following formula. Firstly, the CLIP features are extracted from sampled frames in the video:
\begin{equation}
    f_{video} = \{f_0, f_1, \dots, f_n\} = 
    CLIP(V_0, V_1, \dots, V_n)
\end{equation}

Next, the extracted features \( f_{video} \) are processed by the classification head, for which we experimented with three different variants:\\

1. Average Pooling:
\begin{equation}
    f_{avg} = \frac{1}{n} \sum_{i=0}^{n} f_i
\end{equation}
\begin{equation}
    \hat{y} = \text{Linear}(f_{avg})
\end{equation}

2. Maximum Pooling:
\begin{equation}
    f_{max} = \max(f_0, f_1, \dots, f_n)
\end{equation}
\begin{equation}
    \hat{y} = \text{Linear}(f_{max})
\end{equation}

3. Multi-Layer Perceptron (MLP):
\begin{equation}
    \hat{y} = \text{MLP}(f_{video})
\end{equation}

where \( \hat{y} \) represents the prediction. These different designs aim to capture various temporal dynamics and provide a comparative analysis of their effectiveness in recognizing toddler actions.

\subsection{Skeleton-based Methods}
Skeleton-based methods for human action recognition utilize the structural information from human joints and bones. Given the human pose data generated by upstream pose estimation method, these action recognizers focus on capturing the dynamics and interrelationships of various body parts over time, thereby effectively understanding complex human motions.

\subsubsection{ST-GCN.}
The Spatio-Temporal Graph Convolutional Network (ST-GCN) \cite{stgcn} is a graph convolution-based action recognition model in the field of skeleton-based action recognition. It extends traditional convolutional neural networks (CNNs) to graph structures, where human skeletons are represented as graphs with joints as nodes and bones as edges. ST-GCN effectively captures spatial and temporal patterns by applying graph convolutions in both spatial and temporal dimensions. This allows the model to learn hierarchical features from the skeletal data, making it adept at recognizing intricate human actions.

\subsubsection{PoseConv3D.}
PoseConv3D \cite{duan2022revisiting} is a sophisticated approach that integrates 3D convolutions with pose estimation data to perform action recognition. By leveraging 3D convolutions, PoseConv3D can capture spatiotemporal features directly from the sequence of pose estimations. This method benefits from the ability to learn temporal dependencies and spatial relationships simultaneously, resulting in a robust representation of human actions. PoseConv3D is particularly effective in scenarios where precise temporal information is crucial for distinguishing between similar actions.
\begin{figure}[t]
    \centering
    \includegraphics[width=1\linewidth]{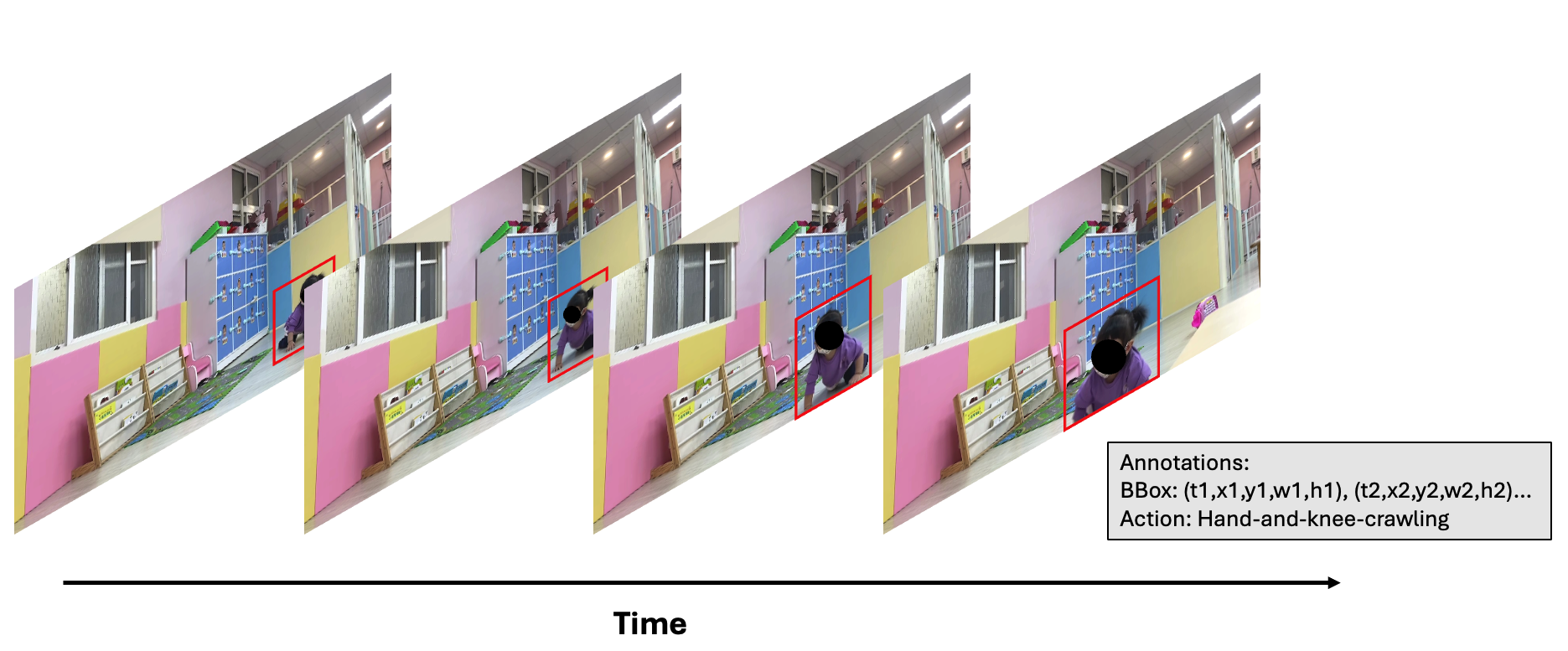}
    \caption{A visualization example of our annotation. Our ToddlerAct dataset's annotation include video-level tracklet's bounding box annotation and the corresponding action. In the situation of multiple toddlers in a video, we include the tracking ID and corresponding action annotation.}
    \label{fig:anno}
\end{figure}
\section{Experiment}
\subsection{Implementation Details}
\subsubsection{CLIP-based Method.}We use the pretrained ViT-B/32 weight from the official implementation of the CLIP model to extract video frame feature $f_{video} = \{f_0, f_1, \dots, f_n\} \in R^{n\times d}$, with $n$ represents the number of sampled frames, which we set to 30 in our experiments. $d$ is the hidden dimension of CLIP. The model was trained using the Cross-Entropy Loss and optimized with the Adam optimizer, with a learning rate set to 0.001 and batch size 128. All the experiments were conducted on a single NVIDIA 4080 GPU. We conduct experiments on ToddlerAct dataset and split the dataset into training and testing set with a 7:3 ratio, with training set containing 389 videos, and testing set containing 167 videos.
\subsubsection{ST-GCN \& PoseConv3D.} To further investigate the current state-of-the-art method's performance on our dataset, we conduct experiments using ST-GCN \cite{stgcn} and PoseConv3D \cite{duan2022revisiting} on our proposed ToddlerAct dataset. ST-GCN and PoseConv3D perform skeleton-based action recognition based on the extracted 2D human keypoints. We trained ST-GCN and PoseConv3D models on our proposed dataset using the tracking result annotations and the extracted human 2D keypoints from HRNet \cite{hrnet} pretrained on COCO \cite{coco}. We used two NVIDIA 2080-Ti GPUs and trained both models for 24 epochs, following the default training setting in PYSKL\cite{pyskl}.

\subsection{Performance}
We conduct experiments evaluating the performance of our method under different settings, including 1. using CLIP zero shot classification on the average image feature from the sampled video feature $f_{video}$, 2. different variants of pooling among the temporal dimension conducted on $f_{video}$ followed by a linear classification layer, 3. a multi-layer perceptron that takes $f_{video}$ as input and generate prediction, 4. state-of-the-art action recognition methods like ST-GCN and PoseConv3D.

For the zero shot CLIP classification, we followed the original CLIP's text prompt template with our action name. E.g. "\textit{A photo of a toddler \{\textbf{action name}\}}", with \textit{\textbf{action name}} taken from the six gross motor action names including hands-and-knees crawling, standing with assistance, walking with assistance, standing alone, and walking alone.

\begin{figure}[t]
    \centering
    \begin{minipage}{0.32\linewidth}
        \centering
        \includegraphics[width=\linewidth]{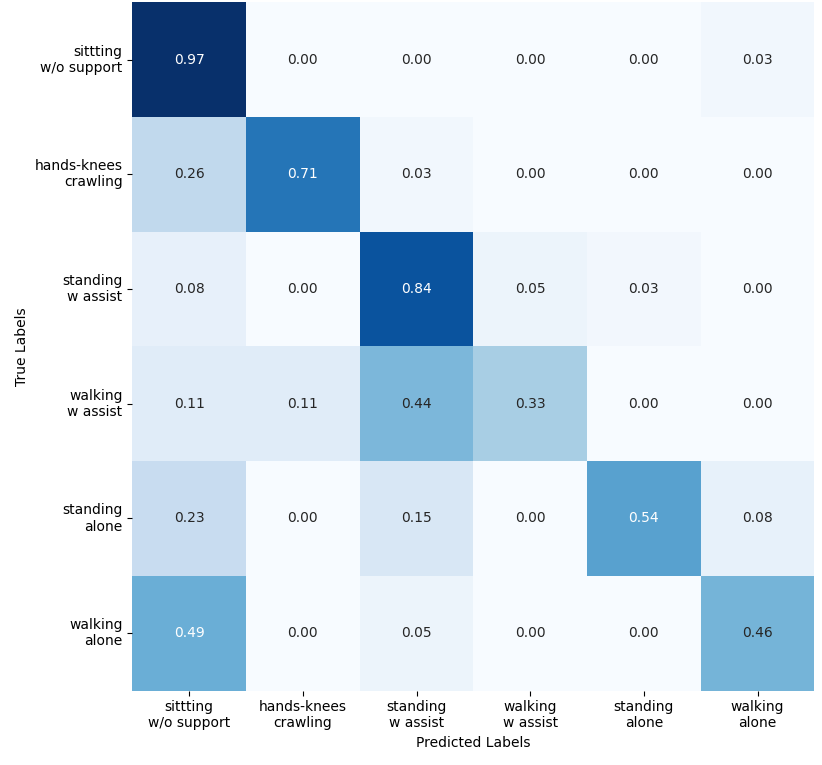}
        \label{fig:confusion1}
    \end{minipage}%
    \begin{minipage}{0.32\linewidth}
        \centering
        \includegraphics[width=\linewidth]{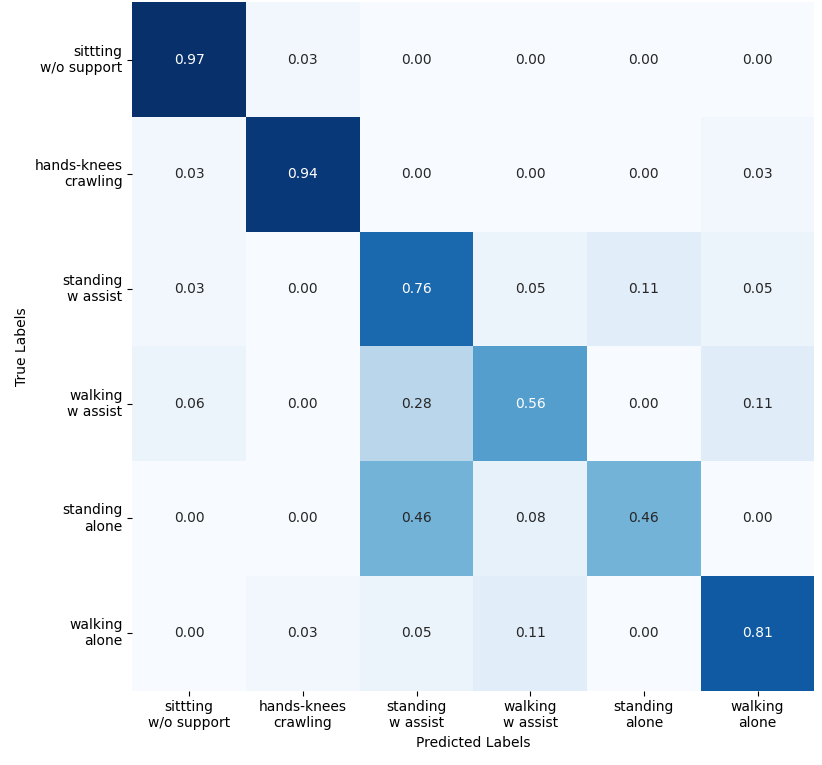}
        \label{fig:confusion2}
    \end{minipage}%
    \begin{minipage}{0.36\linewidth}
        \centering
        \includegraphics[width=\linewidth]{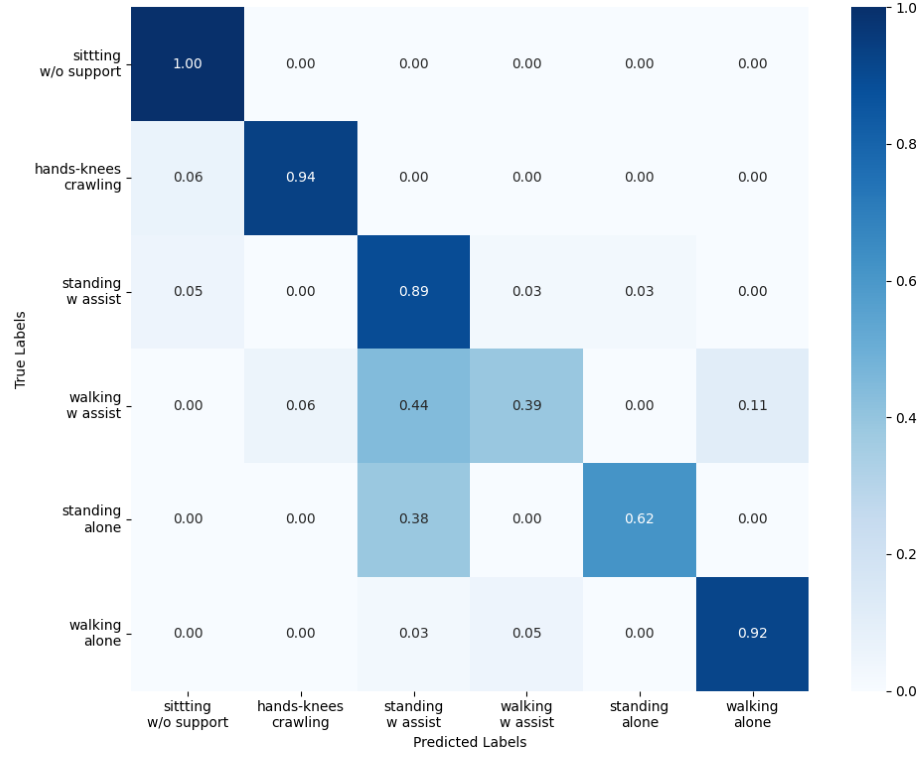}
        \label{fig:confusion3}
    \end{minipage}
    \caption{Confusion matrix from each of the tested baseline, including CLIP \& Multi-layer Perceptron 
 (left), STGCN (center), and PoseConv3D (right).}
    \label{fig:confusion}
\end{figure}

In our experiments, zero shot classification resulted in the lowest classification accuracy, with only 20.5\%. Which might be caused by the information loss during temporal pooling, as well as the CLIP feature is not generalized well on video data with temporal information. The temporal pooling method with linear classifier can perform slightly better, with maximum pooling achieve 25.7\% accuracy and average pooling achieve 41.3\% accuracy. The multi-layer perceptron that takes the whole $f_{video}$ as input without any temporal pooling achieve the highest performance among all the baseline methods, with an action classification accuracy of 67.6\%. We further show the results of the multi-layer perceptron method's confusion matrix in Fig. \ref{fig:confusion}, as shown in Fig. \ref{fig:confusion}, we noticed a much lower accuracy in walking with assistance and standing alone, with only 33.3\% and 53.8\% classification accuracy, respectively. This might be caused by the smaller amount of data of these two actions in our ToddlerAct dataset and thus degrade the performance of on these two actions.

\begin{table*}[h]
\caption{Performance comparison on Top-1 Accuracy of different methods on the ToddlerAct dataset.
}
  \begin{center}
    {\footnotesize{
\begin{tabular}{c|c}
\toprule
Method & Top-1 Acc\\
\toprule
Zero shot Classification & 20.5 \\
Max Pool \& Linear classifier & 25.7 \\
Avg Pool \& Linear classifier & 41.3 \\
Multi-layer Perceptron & 67.6 \\
\bottomrule
\end{tabular}
}}
\end{center}
\label{table:performance}
\end{table*} 


\begin{figure}[t]
    \centering
    \includegraphics[width=1\linewidth, height=100pt]
    {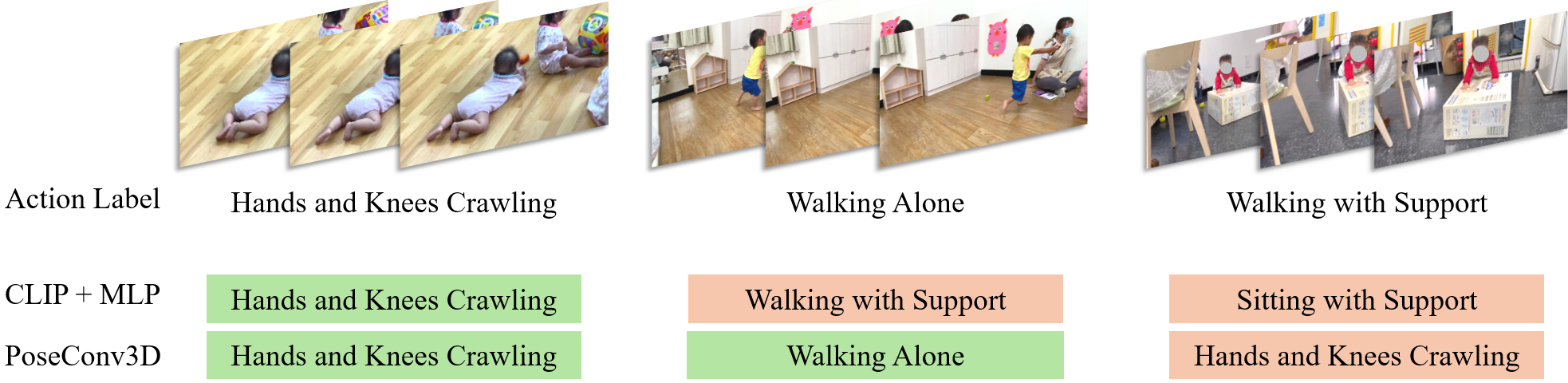}
    \caption{A qualitative comparison between our proposed CLIP + MLP method and the PoseConv3D state-of-the-art method on three video sequences. Both methods correctly assign the action label for the first video on the left ("Hands and Knees Crawling"). For the second video in the middle ("Walking Alone"), only PoseConv3D produces the correct action label. For the third video on the right ("Walking with Support"), neither model correctly recognizes the action correctly due to the heavy occlusion.}

    \label{fig:comparison}
\end{figure}
\begin{table*}[h]
\caption{Performance of different tested methods including CLIP \& MLP, ST-GCN \cite{stgcn}, and PoseConv3D \cite{duan2022revisiting} on the ToddlerAct test set. Stand-A and Walk-A represents standing with assistance and walking with assistance, respectively.}
  \begin{center}
    {\footnotesize{
\begin{tabularx}{\textwidth}{>{\centering\arraybackslash}X|>{\centering\arraybackslash}X|>{\centering\arraybackslash}X|>{\centering\arraybackslash}X|>{\centering\arraybackslash}X|>{\centering\arraybackslash}X|>{\centering\arraybackslash}X|>{\centering\arraybackslash}X}
\toprule
Method & Sit & Crawl & Stand-A & Walk-A & Stand & Walk & All \\
\midrule
CLIP & 96.7 & 70.9 & 83.7 & 33.3 & 53.8 & 45.9 & 67.6\\
ST-GCN & 96.7 & 93.5 & 75.6 & 55.6 & 46.2 & 81.1 & 77.3 \\
PoseC3D & 100.0 & 93.5 & 89.2 & 38.9 & 61.5 & 91.8 & 82.0 \\
\bottomrule
\end{tabularx}
}}
\end{center}
\label{table:posec3d}
\end{table*}
Further experiments on state-of-the-art methods are shown in Table \ref{table:posec3d}. Both ST-GCN and PoseConv3D demonstrated strong performance on the ToddlerAct dataset. Different from the observations of previous works that focus on infant understanding \cite{infact,yang2022unsupervised}, which claims the challenges in infant pose estimation, we found that toddler pose estimation is less challenging compared to infant pose estimation. This is mainly due to toddlers' more mature body development, which makes the human pose estimation model like HRNet \cite{hrnet} pretrained on large-scale human pose estimation dataset also extremely accurate on toddler pose estimation. The accurate pose estimation results further lead to a higher performance of skeleton-based action recognition method over our proposed image-based method.

A qualitative comparison between a simple image-based CLIP+MLP method and the SOTA method, PoseConv3D, is demonstrated in Figure \ref{fig:comparison}. 
Although several recent methods \cite{li2022uniformerv2,wang2024internvideo2} has shown image modality can contain more information and thus resulted in higher performance compared with skeleton-based methods on action recognition task, we believe under the conditions of limited image and video training data, human skeleton can still serves as an important clue to conduct action recognition.
\section{Discussion}
\subsection{Ethical Concerns}
The collection and use of data involving toddlers raise several ethical concerns that must be carefully addressed to ensure the safety, privacy, and well-being of the children involved. Consent from parents or legal guardians is paramount, and they must be fully informed about the nature of the study, the data being collected, and how it will be used. Anonymity and confidentiality of the participants must be strictly maintained, with all identifiable information securely protected. Additionally, it is crucial to implement protocols that ensure the comfort and safety of the children during data collection, minimizing any potential stress or discomfort. Ethical review boards should oversee the study to ensure that all procedures adhere to ethical guidelines and standards.

During the collection process of our ToddlerAct dataset, we make sure to get consent from both parents and the legal guardians for the permission to collect data. The collection of the dataset is also carefully reviewed and approved by the national human research ethics committee. Furthermore, all the toddlers' face will be masked in the released ToddlerAct dataset in order to protect their privacy.

\subsection{Future Work}
We discussed several of our future works in this section. In our proposed ToddlerAct dataset, some of the actions including standing alone and walking with assistance have a smaller amount of data compared with other actions, this leads to a lower classification accuracy for these classes. Collecting more data regarding actions or implement method that can deal with class imbalanced dataset is part of our future direction.

Furthermore, the performance of our the image-based method still has room for improvement compare with skeleton-based method, the CLIP-based method failed in some cases where complicated human object interaction understanding is needed to achieve correct prediction, e.g. walking with assistance needs to identify the toddler is supported by other adult or object. We will seek to some human object interaction method as one of our solution to improve the performance.

Finally, longitudinal studies that track the development of gross motor skills over time can provide valuable insights into the progression of toddler development. Incorporating such longitudinal data into the dataset and analysis can help in understanding the dynamics of motor skill acquisition and identifying early markers of developmental delays.

\section{Conclusion}
In this paper, we propose a toddler action recognition dataset collected from real-world scenarios, aimed at facilitating current research in toddler action recognition and gross motor assessment. Furthermore, we tested and benchmarked multiple different state-of-the-art methods including image-based and skeleton-based as our baselines. We hope that our contributions will provide a valuable resource and benchmark for future research in this domain, encouraging the development of more advanced and accurate models for assessing gross motor skills in toddlers.

\bibliographystyle{splncs04}
\bibliography{ref}

\end{document}


\title{Supplementary Material}

\titlerunning{Exploring Learning-based Motion Models in Multi-Object Tracking}

\author{
Hsiang-Wei Huang\orcidlink{0009-0009-2474-8869} \and
Cheng-Yen Yang\orcidlink{0009-0004-2631-6756} \and
Wenhao Chai\orcidlink{0000-0003-2611-0008} \and 
\\
Zhongyu Jiang\orcidlink{0000-1234-1234-5555} \and
Jenq-Neng Hwang\orcidlink{0000-0002-8877-2421}
}

\authorrunning{H.-W. Huang et al.}

\institute{
Department of Electrical \& Computer Engineering \\
University of Washington, Seattle WA, USA \\
\email{\{hwhuang, cycyang, wchai, zyjiang, hwang\}@uw.edu}\\}

\maketitle

\input{tex/7_supp}

%
%

\clearpage
\bibliographystyle{splncs04}
\bibliography{ref}